\documentclass[10pt,twocolumn,letterpaper]{article}

\usepackage{iccv}
\usepackage{times}
\usepackage{epsfig}
\usepackage{graphicx}
\usepackage{amsmath}
\usepackage{amssymb}

\usepackage{amsfonts,amssymb} 

\usepackage{booktabs}
\usepackage[numbers,sort&compress]{natbib}
\usepackage{color}
\usepackage[normalem]{ulem}

\usepackage{amsmath}
\usepackage{algorithm}
\usepackage[noend]{algpseudocode}

\usepackage{float}
\usepackage{multirow}
\usepackage{epsfig}

\usepackage{balance}

\usepackage[pagebackref=true,breaklinks=true,letterpaper=true,colorlinks,bookmarks=false]{hyperref}

\iccvfinalcopy 

\def\iccvPaperID{****} 

\ificcvfinal\fi

\begin{document}

\title{Visual Semantic Reasoning for Image-Text Matching}

\author{
   Kunpeng Li$^{1}$, Yulun Zhang$^{1}$, Kai Li$^{1}$, Yuanyuan Li$^{1}$ and Yun Fu$^{1,2}$\\
   $^1$Department of Electrical and Computer Engineering, Northeastern University, Boston, MA\\
   $^2$Khoury College of Computer Science, Northeastern University, Boston, MA\\
}

\maketitle
\ificcvfinal\thispagestyle{empty}\fi

\begin{abstract}

Image-text matching has been a hot research topic bridging the vision and language areas. It remains challenging because the current representation of image usually lacks global semantic concepts as in its corresponding text caption. To address this issue, we propose a simple and interpretable reasoning model to generate visual representation that captures key objects and semantic concepts of a scene. Specifically, we first build up connections between image regions and perform reasoning with Graph Convolutional Networks to generate features with semantic relationships. Then, we propose to use the gate and memory mechanism to perform global semantic reasoning on these relationship-enhanced features, select the discriminative information and gradually generate the representation for the whole scene. Experiments validate that our method achieves a new state-of-the-art for the image-text matching on MS-COCO~\cite{lin2014microsoft} and Flickr30K~\cite{young2014image} datasets. It outperforms the current best method by 6.8\% relatively for image retrieval and 4.8\% relatively for caption retrieval on MS-COCO (Recall@1 using 1K test set). On Flickr30K, our model improves image retrieval by 12.6\% relatively and caption retrieval by 5.8\% relatively (Recall@1). Our code is available at \url{https://github.com/KunpengLi1994/VSRN}.



\end{abstract}

\section{Introduction}


Vision and language are two important aspects of human intelligence to understand the real world. A large amount of research \cite{lee2018stacked,faghri2017vse,gu2019scene} has been done to bridge these two modalities. Image-text matching is one of the fundamental topics in this field, which refers to measuring the visual-semantic similarity between a sentence and an image. It has been widely adopted to various applications such as the retrieval of text descriptions from image queries or image search for given sentences. 

\begin{figure}
\centering
\includegraphics[width=1.0\linewidth]{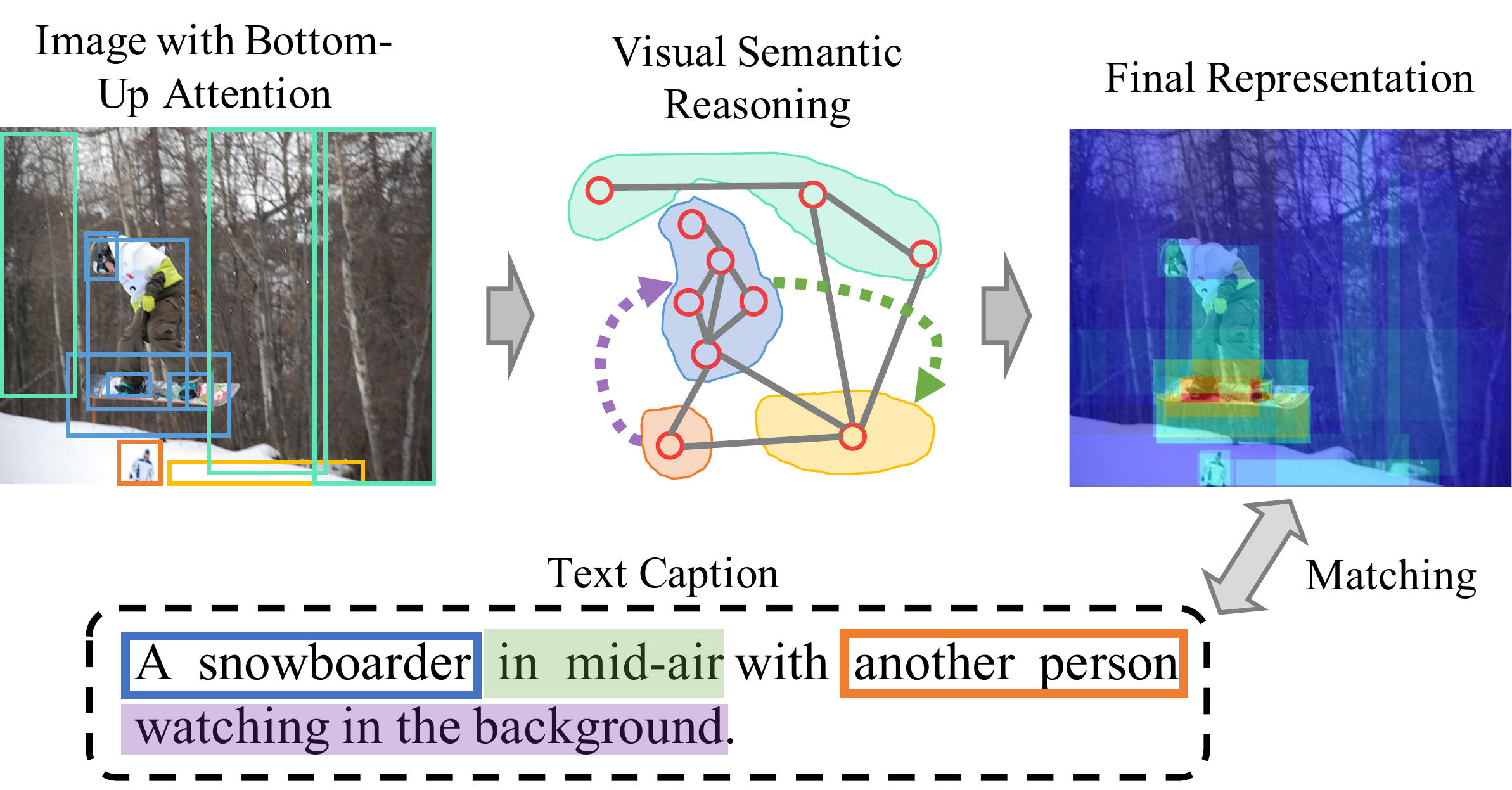} 
\caption{The proposed Visual Semantic Reasoning Network (VSRN) performs reasoning on the image regions to generate representation for an image. The representation captures key objects (boxes in the caption) and semantic concepts (highlight parts in the caption) of a scene as in the corresponding text caption. }
\label{fig:teaser} 
\end{figure}

Although a lot of progress has been achieved in this area, it is still a challenge problem due to the huge visual semantic discrepancy. When people describe what they see in the picture using natural language, it can be observed that the descriptions will not only include the objects, salient stuff, but also will organize their interactions, relative positions and other high-level semantic concepts (such as ``in mid-air'' and ``watching in the background'' in the Figure~\ref{fig:teaser}). Visual reasoning about objects and semantics is crucial for humans during this process. However, the current existing visual-text matching systems lack such kind of reasoning mechanism. Most of them \cite{faghri2017vse} represent concepts in an image by Convolutional Neural Network (CNN) features extracted by convolutions with a specific receptive field, which only perform local pixel-level analysis. It is hard for them to recognize the high-level semantic concepts. More recently, \cite{lee2018stacked} make use of region-level features from object detectors and discover alignments between image regions and words. Although grasping some local semantic concepts within regions including multiple objects, these methods still lack the global reasoning mechanism that allows information communication between regions farther away.



To address this issue, we propose Visual Semantic Reasoning Network (VSRN) to generate visual representation that captures both objects and their semantic relationships. We start from identifying salient regions in images by following \cite{lee2018stacked,anderson2018bottom}. In this way, salient region detection at stuff/object level can be analogized to the bottom-up attention that is consistent with human vision system~\cite{katsuki2014bottom}. Practically, the bottom-up attention module is implemented using Faster R-CNN~\cite{ren2015faster}. We then build up connections between these salient regions and perform reasoning with Graph Convolutional Networks (GCN) \cite{kipf2017semi} to generate features with semantic relationships.


Different image regions and semantic relationships would have different contributions for inferring the image-text similarity and some of them are even redundant. Therefore, we further take a step to attend important ones when generating the final representation for the whole image. We propose to use the gate and memory mechanism \cite{chung2014empirical} to perform global semantic reasoning on these relationship-enhanced features, select the discriminative information and gradually grow representation for the whole scene. This reasoning process is conducted on a graph topology and considers both local, global semantic correlations. The final image representation captures more key semantic concepts than those from existing methods that lack a reasoning mechanism, therefore, can help to achieve better image-text matching performance. 


In addition to quantitative evaluation of our model on standard benchmarks, we also design an interpretation method to analyze what has been learned inside the reasoning model. Correlations between the final image representation and each region feature are visualized in an attention format. As shown in Figure \ref{fig:teaser}, we find the learned image representation has high response at these regions that include key semantic concepts. 


To sum up, our main contributions are: (a) We propose a simple and interpretable reasoning model VSRN to generate enhanced visual representations by region relationship reasoning and global semantic reasoning. (b) We design an interpretation method to visualize and validate that the generated image representation can capture key objects and semantic concepts of a scene, so that it can be better aligned with the corresponding text caption. (c) The proposed VSRN achieves a new state-of-the-art for the image-text matching on MS-COCO~\cite{lin2014microsoft} and Flickr30K~\cite{young2014image} datasets. Our VSRN outperforms the current best method SCAN \cite{lee2018stacked} by 6.8\% relatively for image retrieval and 4.8\% relatively for caption retrieval on MS-COCO (Recall@1 using 1K test set). On Flickr30K, our model improves image retrieval by 12.6\% relatively and caption retrieval by 5.8\% relatively (Recall@1).


\section{Related Work}


\textbf{Image-Text Matching.}
Our work is related to existing methods proposed for image-text matching, where the key issue is measuring the visual-semantic similarity between a text and an image. Learning a common space where text and image feature vectors are comparable is a typical solution for this task. Frome et al. \cite{frome2013devise} propose a feature embedding framework that uses Skip-Gram and CNN to extract feature representations for cross-modal. Then a ranking loss is adopted to encourage the distance between the mismatched
image-text pair is larger than that between the matched pair. Kiros et al. \cite{kiros2014unifying} use a similar framework and adopt LSTM \cite{hochreiter1997long} instead of Skip-Gram for the learning of text representations. Vendrov et al. \cite{vendrov2015order} design a new objective function that encourages the order structure of visual semantic can be preserved hierarchy. Faghri et al. \cite{faghri2017vse} focus more on hard negatives and obtain good improvement using a triplet loss. Gu et al. \cite{gu2018look} further improve the learning of cross-view feature embedding by incorporating generative objectives. Our work also belongs to this direction of learning joint space for image and sentence with an emphasis on improving image representations. 



\textbf{Attention Mechanism.}
Our work is also inspired by bottom-up attention mechanism and recent image-text matching methods based on it. Bottom-up attention \cite{katsuki2014bottom} refers to salient region detection at stuff/object level can be analogized to the spontaneous bottom-up attention that is consistent with human vision system \cite{katsuki2014bottom,li2019gain,li2018tell,li2019attnbn,li2018support}. Similar observation has motivated other existing work. In \cite{karpathy2015deep}, R-CNN \cite{girshick2014rich} is adopted to detect and encode image regions at object level. Image-text similarity is then obtained by aggregating all word-region pairs similarity scores. Huang et al. \cite{huang2018learning} train a multi-label CNN to classify each image region into multi-labels of objects and semantic relations, so that the improved image representation can capture semantic concepts within the local region. Lee et al. \cite{lee2018stacked} further propose an attention model towards attending key words and image regions for predicting the text-image similarity. Following them, we also start from bottom-up region features of an image. However, to the best of our knowledge, no study has attempted to incorporate global spatial or semantic reasoning when learning visual representations for image-text matching.

\begin{figure*}
\centering
\includegraphics[width=1.0\linewidth]{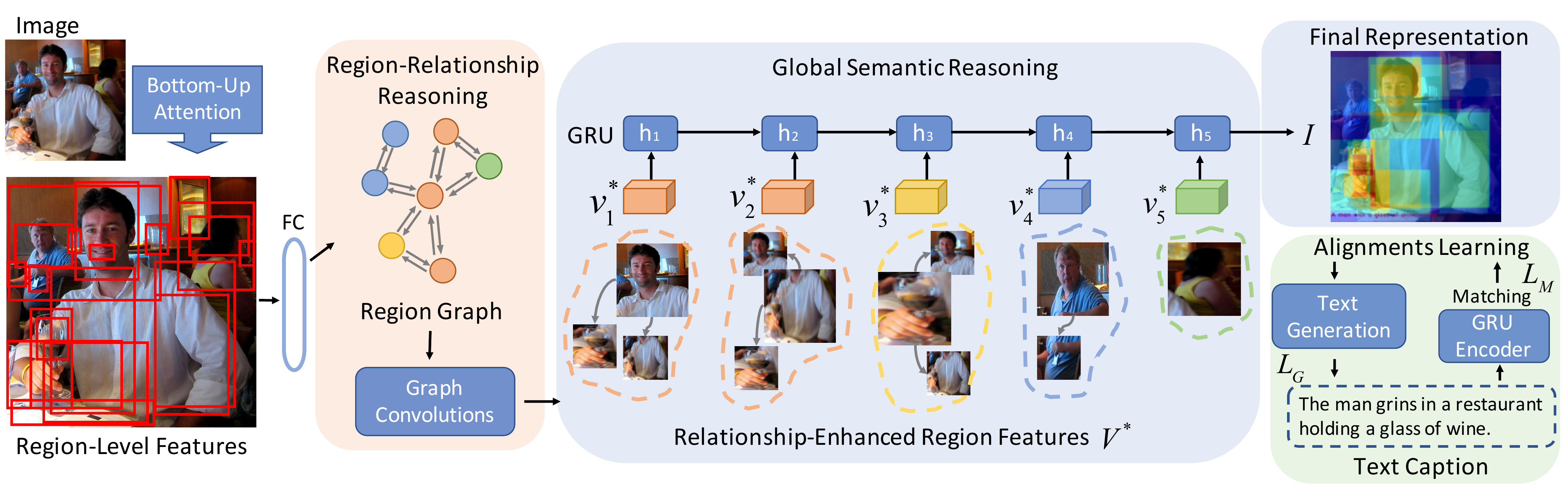} 
\caption{ An overview of the proposed Visual Semantic Reasoning Network (VSRN). Based on salient image regions from bottom-up attention (Sec.~\ref{sc:bottom_up}), VSRN first performs region relationship reasoning on these regions using GCN to generate features with semantic relationships (Sec.~\ref{sc:region_relation_R}). Then VSRN takes use of the gate and memory mechanism to perform global semantic reasoning on the relationship enhanced features, select the discriminative information and gradually generate the representation for the whole scene (Sec.~\ref{sc:global_context_R}). The whole model is trained with joint optimization of matching and sentence generation (Sec.~\ref{sc:alignment}). The attention of the representation (top right) is obtained by calculating correlations between the final image representation and each region feature (Sec.~\ref{sc:visualization}). }




\label{fig:VSRN_model} 
\end{figure*}


\textbf{Relational Reasoning Methods.}
Symbolic approaches \cite{newell1980physical} are the earliest form of reasoning in artificial intelligence. In these methods, relations between symbols are represented by the form of logic and mathematics, reasoning happens by abduction and deduction \cite{hobbs1993interpretation} etc. However, in order to make these systems can be used practically, symbols need to be grounded in advance. More recent methods, such as path ranking algorithm \cite{lao2011random}, perform reasoning on structured knowledge bases by taking use of statistical learning to extract effective patterns. As an active research area, graph-based methods \cite{zhang2019mst} have been very popular in recent years and shown to be an efficient way of relation reasoning. Graph Convolution Networks (GCN) \cite{kipf2017semi} are proposed for semi-supervised classification. Yao et al. \cite{yao2018exploring} train a visual relationship detection model on Visual Genome dataset \cite{krishna2017visual} and use a GCN-based encoder to encode the detected relationship information into an image captioning framework. Yang et al. \cite{yang2018visual} utilize GCNs to incorporate the prior knowledge into a deep reinforcement learning framework improve semantic navigation in unseen scenes and towards novel objects. We also adopt the reasoning power of graph convolutions to obtain image region features enhanced with semantic relationship. But we do not need extra database to build the relation graph (e.g. \cite{yao2018exploring} needs to train the relationship detection model on Visual Genome). Beyond this, we further perform global semantic reasoning on these relationship-enhanced features, so that the final image representation can capture key objects and semantic concepts of a scene.

\section{Learning Alignments with Visual Semantic Reasoning}

We describe the detail structure of the Visual Semantic Reasoning Network (VSRN) for image-text matching in this section. Our goal is to infer the similarity between a full sentence and a whole image by mapping image regions and the text descriptions into a common embedding space. For the image part, we begin with image regions and their features generated by the bottom-up attention model~\cite{anderson2018bottom} (Sec.~\ref{sc:bottom_up}). VSRN first builds up connections between these image regions and do reasoning using Graph Convolutional Networks (GCN) to generate features with semantic relationship information (Sec.~\ref{sc:region_relation_R}). Then, we do global semantic reasoning on these relationship-enhanced features to select the discriminative information and filter out unimportant one to generate the final representation for the whole image (Sec.~\ref{sc:global_context_R}). For the text caption part, we learn a representation for the sentence using RNNs. Finally, the whole model is trained with joint optimization of image-sentence matching and sentence generation (Sec.~\ref{sc:alignment}). 

\subsection{Image Representation by Bottom-Up Attention }\label{sc:bottom_up}


Taking the advantage of bottom-up attention~\cite{anderson2018bottom}, each image can be represented by a set of features $V = \{ {v_1},...,{v_k}\} ,{v_i} \in {\mathbb{R}^{{D}}}$, such that each feature $v_{i}$ encodes an object or a salient region in this image. Following~\cite{anderson2018bottom,lee2018stacked}, we implement the bottom-up attention with a Faster R-CNN~\cite{ren2015faster} model using ResNet-101~\cite{he2016deep} as the backbone. It is pre-trained on the Visual Genomes dataset~\cite{krishna2017visual} by~\cite{anderson2018bottom}. The model is trained to predict instance classes and attribute classes instead of the object classes, so that it can help learn feature representations with rich semantic meaning. Specifically, instance classes include objects and salient stuff which is hard to recognize. For example, attributes like ``furry'' and stuff like ``building'', ``grass'' and ``sky''. The model's final output is used and non-maximum suppression for each class is operated with an IoU threshold of 0.7. We then set a confidence threshold of 0.3 and select all image regions where any class detection probability is larger than this threshold. The top 36 ROIs with the highest class detection confidence scores are selected. All these thresholds are set as same as~\cite{anderson2018bottom,lee2018stacked}. For each selected region $i$, we extract features after the average pooling layer, resulting in $f_{i}$ with 2048 dimensions. A fully-connect layer is then applied to transform $f_{i}$ to a $D$-dimensional embedding using the following equation:

\begin{equation}
\label{eq:img_embedding}
{v_i} = {W_f}{f_i} + {b_f}.
\end{equation}

Then $V = \{ {v_1},...,{v_k}\} ,{v_i} \in {\mathbb{R}^{{D}}}$ is constructed to represent each image, where $v_{i}$ encodes an object or salient region in this image.

\subsection{Region Relationship Reasoning}\label{sc:region_relation_R}


Inspired by recent advances in deep learning based visual reasoning~\cite{santoro2017simple,chen2018iterative,zhou2018temporal}, we build up a region relationship reasoning model to enhance the region-based representation by considering the semantic correlation between image regions. Specifically, we measure the pairwise affinity between image regions in an embedding space to construct their relationship using Eq.~\ref{eq:affinity}.

\begin{equation}
\label{eq:affinity}
R({v_i},{v_j}) = \varphi {({v_i})^T}\phi ({v_j}),
\end{equation}
where $\varphi ({v_i}) = {W_\varphi }{v_i}$ and $\phi ({v_j}) = W_\phi {v_j}$ are two embeddings. The weight parameters $W_\varphi$ and $W_\phi$ can be learned via back propagation.

Then a fully-connected relationship graph $G_r = (V,E)$, where $V$ is the set of detected regions and edge set $E$ is described by the affinity matrix $R$. $R$ is obtained by calculating the affinity edge of each pair of regions using Eq.~\ref{eq:affinity}. That means there will be an edge with high affinity score connecting two image regions if they have strong semantic relationships and are highly correlated. 


We apply the Graph Convolutional Networks (GCN) \cite{kipf2017semi} to perform reasoning on this fully-connected graph. Response of each node is computed based on its neighbors defined by the graph relations. We add residual connections to the original GCN as follows: 

\begin{equation}
\label{eq:GCN}
{V^*} = {W_r}(RV{W_g}) + V,
\end{equation}
where $W_g$ is the weight matrix of the GCN layer with dimension of $D \times D$. $W_r$ is the weight matrix of residual structure. $R$ is the affinity matrix with shape of $k \times k$. We follow the routine to row-wise normalize the affinity matrix $R$. The output $V^* = \{ {v_1^*},...,{v_k^*}\} ,{v_i^*} \in {\mathbb{R}^{{D}}}$ is the relationship enhanced representation for image region nodes.

\subsection{Global Semantic Reasoning}\label{sc:global_context_R} 


Based on region features with relationship information, we further do global semantic reasoning to select the discriminative information and filter out unimportant one to obtain the final representation for the whole image. Specifically, we perform this reasoning by putting the sequence of region features $V^* = \{ {v_1^*},...,{v_k^*}\} ,{v_i^*} \in {\mathbb{R}^{{D}}}$, one by one into GRUs \cite{chung2014empirical}. The description of the whole scene will gradually grow and update in the memory cell (hidden state) $m_i$ during this reasoning process. 

At each reasoning step $i$, an update gate $z_i$ analyzes the current input region feature $v_i^*$ and the description of the whole scene at last step $m_{i-1}$ to decide how much the unit updates its memory cell. The update gate is calculated by:

\begin{equation}
\label{eq:update_gate}
z_i = {\sigma _z}({W_z}v_i^* + {U_z}{m_{i - 1}} + {b_z}),
\end{equation}
where ${\sigma _z}$ is a sigmoid activation function. $W_z$, $U_z$ and $b_z$ are weights and bias. 

The new added content helping grow the description of the whole scene is computed as follows:

\begin{equation}
\label{eq:new_content}
{\tilde m_i} = {\sigma _m}({W_m}v_i^* + {U_z}({r_i} \circ {m_{i - 1}}) + {b_m}),
\end{equation}
where ${\sigma _m}$ is a $\tanh$ activation function. $W_m$, $U_m$ and $b_m$ are weights and bias. $\circ$ is an element-wise multiplication. ${r_i}$ is the reset gate that decides what content to forget based on the reasoning between $v_i^*$ and $m_{i-1}$. ${r_i}$ is computed similarly to the update gate as:

\begin{equation}
\label{eq:reset_gate}
r_i = {\sigma _r}({W_r}v_i^* + {U_r}{m_{i - 1}} + {b_r}),
\end{equation}
where ${\sigma _r}$ is a sigmoid activation function. $W_z$, $U_z$ and $b_z$ are weights and bias. 

Then the description of the whole scene $m_{i}$ at the current step is a linear interpolation using update gate $z_i$ between the previous description $m_{i-1}$ and the new content ${\tilde m_i}$:

\begin{equation}
\label{eq:new_memory}
{m_i} = (1 - {z_i}) \circ {m_{i - 1}} + {z_i} \circ {\tilde m_i},
\end{equation}
where $\circ$ is an element-wise multiplication. Since each $v_i^*$ includes global relationship information, update of ${m_i}$ is actually based on reasoning on a graph topology, which considers both current local region and global semantic correlations. We take the memory cell $m_{k}$ at the end of the sequence $V^*$ as the final representation $I$ for the whole image, where $k$ is the length of $V^*$.



\subsection{Learning Alignments by Joint Matching and Generation}\label{sc:alignment}

To connect vision and language domains, we use a GRU-based text encoder  \cite{chung2014empirical,faghri2017vse} to map the text caption to the same $D$-dimensional semantic vector space ${C} \in {\mathbb{R}^{{D}}}$ as the image representation $I$, which considers semantic context in the sentence. Then we jointly optimize matching and generation to learn the alignments between ${C}$ and $I$.

For the matching part, we adopt a hinge-based triplet ranking loss \cite{karpathy2015deep,faghri2017vse,lee2018stacked} with emphasis on hard negatives \cite{faghri2017vse}, i.e., the negatives closest to each training query. We define the loss as:


\begin{equation}
\label{eq:loss_match}
\begin{split}
{L_M} = & {{{[\alpha  - S(I,C) + S(I,\hat C)]}_ + }} + \\
& {{{[\alpha  - S(I,C) + S(\hat I,C)]}_ + }},
\end{split}
\end{equation}
where $\alpha$ serves as a margin parameter. ${[x]_ + } \equiv \max (x,0)$. This hinge loss comprises two terms, one with $I$ and one with $C$ as queries. $S( \cdot )$ is the similarity function in the joint embedding space. We use the usual inner product as $S( \cdot ) $in our experiments. $\hat I = \arg {\max _{j \ne I}}S(j,C)$ and $\hat C = \arg {\max _{d \ne C}}S(I,d)$ are the hardest negatives for a positive pair (I, T). For computational efficiency, instead of finding the hardest negatives in the entire training set, we find them within each mini-batch.

For the generation part, the learned visual representation should also has the ability to generate sentences that are close to the ground-truth captions. Specifically, we use a sequence to sequence model with attention mechanism \cite{venugopalan2015sequence} to achieve this. We maximize the log-likelihood of the predicted output sentence. The loss function is defined as:


\begin{equation}
\label{eq:loss_generation}
{L_G} =  - \sum\limits_{t = 1}^l {\log p({y_t}|{y_{t - 1}},{V^*};\theta )},
\end{equation}
where $l$ is the length of output word sequence $Y = ({y_1},...,{y_l})$. $\theta$ is the parameter of the sequence to sequence model.  

Our final loss function is defined as follows to perform joint optimization of the two objectives.

\begin{equation}
\label{eq:final_loss}
 L = {L_M} + {L_G}.
\end{equation}

\section{Experiments}

To evaluate the effectiveness of the proposed Visual Semantic Reasoning Network (VSRN), we perform experiments in terms of sentence retrieval (image query) and image retrieval (sentence query) on two publicly available datasets. Ablation studies are conducted to investigate each component of our model. We also compare with recent state-of-the-art methods on this task.

\subsection{Datasets and Protocols}

We evaluate our method on the Microsoft COCO dataset~\cite{lin2014microsoft} and the Flickr30K dataset~\cite{young2014image}. MS-COCO includes 123,287 images, and each image is annotated with 5 text descriptions. We follow the splits of \cite{karpathy2015deep,gu2018look,faghri2017vse,lee2018stacked} for MSCOCO, which contains 113,287 images for training, 5,000 images for validation and 5000 images for testing. Each image comes with 5 captions. The final results are obtained by averaging the results from 5 folds of 1K test images or testing on the full 5K test images. Flickr30K consists of 31783 images collected from the Flickr website. Each image is accompanied with 5 human annotated text descriptions. We use the standard training, validation and testing splits~\cite{karpathy2015deep}, which contain 28,000 images, 1000 images and 1000 images respectively. For evaluation matrix, as is common in information retrieval, we measure the performance by recall at K (R@K) defined as the fraction of queries for which the correct item is retrieved in the closest K points to the query.

\subsection{Implementation Details}
We set the word embedding size to 300 and the dimension of the joint embedding space $D$ to 2048. We follow the same setting as \cite{lee2018stacked,anderson2018bottom} to set details of visual bottom-up attention model. The order of regions for GRU-based global semantic reasoning (Sec.~\ref{sc:global_context_R}) is determined by the descending order of their class detection confidence scores that are generated by the bottom-up attention detector. For the training of VSRN, we use the Adam optimizer \cite{kingma2014adam} to train the model with 30 epochs. We start training with learning rate 0.0002 for 15 epochs, and then lower the learning rate to 0.00002 for the rest 15 epochs. We set the margin $\alpha $ in Eq.~\ref{eq:loss_match} to 0.2 for all experiments. We use a mini-batch size of 128. For evaluation on the test set, we tackle over-fitting by choosing the snapshot of the model that performs best on the validation set. The best snapshot is selected based on the sum of the recalls on the validation set.

\subsection{Comparisons With the State-of-the-art}

\begin{table}
\centering
\scalebox{0.77}{
\begin{tabular}{l|ccc|ccc}
\hline
\multirow{2}{*}{Methods} & \multicolumn{3}{c|}{Caption Retrieval} & \multicolumn{3}{c}{Image Retrieval} \\
&  R@$1$ & R@$5$ & R@$10$ & R@$1$ & R@$5$ & R@$10$ \\      
\hline
\multicolumn{1}{l}{(R-CNN, AlexNet) } \\
DVSA$_{\rm{CVPR'15}}$ \cite{karpathy2015deep}  & 38.4 & 69.9 & 80.5 & 27.4  & 60.2 & 74.8 \\
HMlstm$_{\rm{ICCV'17}}$ \cite{niu2017hierarchical}  & 43.9 & - & 87.8 & 36.1 & - & 86.7  \\
\hline
\multicolumn{1}{l}{(VGG)} \\
FV$_{\rm{CVPR'15}}$ \cite{klein2015associating}  & 39.4 & 67.9 & 80.9 & 25.1 & 59.8 & 76.6 \\
OEM$_{\rm{ICLR'16}}$ \cite{vendrov2015order}  & 46.7 & - & 88.9 & 37.9 & - & 85.9 \\
VQA$_{\rm{ECCV'16}}$ \cite{lin2016leveraging}  & 50.5 & 80.1 & 89.7 & 37.0 & 70.9 & 82.9 \\
SMlstm$_{\rm{CVPR'17}}$ \cite{huang2017instance}  &  53.2 & 83.1 & 91.5 & 40.7 & 75.8 & 87.4  \\
2WayN$_{\rm{CVPR'17}}$ \cite{eisenschtat2017linking}  & 55.8 & 75.2 & - & 39.7 & 63.3 &  - \\
\hline
\multicolumn{1}{l}{(ResNet)} \\
RRF$_{\rm{ICCV'17}}$ \cite{liu2017learning}  & 56.4 & 85.3 & 91.5 & 43.9 & 78.1 & 88.6 \\
VSE++$_{\rm{BMVC'18}}$ \cite{faghri2017vse}  &  64.6 & 89.1 & 95.7 & 52.0 & 83.1 & 92.0  \\
GXN$_{\rm{CVPR'18}}$ \cite{gu2018look}  & 68.5 & - & 97.9 & 56.6 & - & 94.5  \\
SCO$_{\rm{CVPR'18}}$ \cite{huang2018learning}  & 69.9 & 92.9 & 97.5 & 56.7 &  87.5 & 94.8\\
\hline
\multicolumn{3}{l}{(Faster R-CNN, ResNet) } \\
SCAN$_{\rm{ECCV'18}}$ \cite{lee2018stacked}  & 72.7 & 94.8  & \textbf{98.4} & 58.8 & 88.4 & 94.8 \\  
VSRN (ours)  & \textbf{76.2} & \textbf{94.8}  & 98.2 & \textbf{62.8} & \textbf{89.7} & \textbf{95.1}  \\
\hline
\end{tabular}}
\vspace{1mm}
\caption{Quantitative evaluation results of the image-to-text (caption) retrieval and text-to-image (image) retrieval on MS-COCO 1K test set in terms of Recall@K (R@K).}
\label{tb:coco_results_1K}
\end{table}

\begin{table}
\centering
\scalebox{0.77}{
\begin{tabular}{l|ccc|ccc}
\hline
\multirow{2}{*}{Methods} & \multicolumn{3}{c|}{Caption Retrieval} & \multicolumn{3}{c}{Image Retrieval} \\
&  R@$1$ & R@$5$ & R@$10$ & R@$1$ & R@$5$ & R@$10$ \\      
\hline
\multicolumn{1}{l}{(R-CNN, AlexNet) } \\
DVSA$_{\rm{CVPR'15}}$ \cite{karpathy2015deep}  & 11.8 & 32.5 & 45.4 & 8.9  & 24.9 & 36.3 \\
\hline
\multicolumn{1}{l}{(VGG)} \\
FV$_{\rm{CVPR'15}}$ \cite{klein2015associating}  & 17.3 & 39.0 & 50.2 & 10.8 & 28.3 & 40.1 \\
VQA$_{\rm{ECCV'16}}$ \cite{lin2016leveraging}  & 23.5 & 50.7 & 63.6 & 16.7 & 40.5 & 53.8 \\
OEM$_{\rm{ICLR'16}}$ \cite{vendrov2015order}  & 23.3 & - & 84.7 &  31.7 & - & 74.6 \\
\hline
\multicolumn{1}{l}{(ResNet)} \\
VSE++$_{\rm{BMVC'18}}$ \cite{faghri2017vse}  & 41.3 & 69.2 & 81.2 & 30.3 & 59.1 & 72.4 \\
GXN$_{\rm{CVPR'18}}$ \cite{gu2018look}  & 42.0 & - & 84.7 & 31.7 & - & 74.6 \\
SCO$_{\rm{CVPR'18}}$ \cite{huang2018learning}  & 42.8 & 72.3 & 83.0 & 33.1 & 62.9 & 75.5 \\
\hline
\multicolumn{3}{l}{(Faster R-CNN, ResNet) } \\
SCAN$_{\rm{ECCV'18}}$ \cite{lee2018stacked}  & 50.4 & \textbf{82.2} & \textbf{90.0} & 38.6 & 69.3 & 80.4\\  
VSRN (ours)  & \textbf{53.0} & 81.1  & 89.4 & \textbf{40.5} & \textbf{70.6} & \textbf{81.1} \\
\hline
\end{tabular}}
\vspace{1mm}
\caption{Quantitative evaluation results of the image-to-text (caption) retrieval and text-to-image (image) retrieval on MS-COCO 5K test set in terms of Recall@K (R@K).}
\label{tb:coco_results_5K}
\end{table}

\textbf{Results on MS-COCO.}
Quantitative results on MS-COCO 1K test set are shown in Table~\ref{tb:coco_results_1K}, where the proposed VSRN outperforms recent methods with a large gap of R@1. Following the common protocol \cite{lee2018stacked,huang2018learning,faghri2017vse}, the results are obtained by averaging over 5 folds of 1K test images. When comparing with the current best method SCAN~\cite{lee2018stacked}, we follow the same strategy~\cite{lee2018stacked} to combine results from two trained VSRN models by averaging their predicted similarity scores. Our VSRN improves 4.8\% on caption retrieval (R@1) and 6.8\% on image retrieval (R@1) relatively. In Table~\ref{tb:coco_results_5K}, we also report results on MS-COCO 5K test set by testing on the full 5K test images and their captions. From the table, we can observe that the overall results by all the methods are lower than the first protocol. It probably results from the existence of more distracters for a given query in such a larger target set. Among all methods, the proposed VSRN still achieves the best performance, which again demonstrates its effectiveness. It improves upon the current state-of-the-art, SCAN with 5.2\% on the sentence retrieval (R@1) and 4.9\% on the image retrieval (R@1) relatively.

\textbf{Results on Flickr30K.} 
We show experimental results of VSRN on Flickr30K dataset and comparisons with the current state-of-the-art methods in Table~\ref{tb:flicker_results}. We also list the network backbones used for visual feature extraction, such as R-CNN, VGG, ResNet, Faster R-CNN. From the results, we find the proposed VSRN outperforms all state-of-the-art methods, especially for Recall@1. When compared with SCAN~\cite{lee2018stacked} that uses the same feature extraction backbones with us, our VSRN improves 5.8\% on caption retrieval (R@1) and 12.6\% on image retrieval(R@1) relatively (following the same strategy~\cite{lee2018stacked} of averaging predicted similarity scores of two trained models). SCAN tries to discover the full latent alignments between possible pairs of regions and words, and builds up an attention model to focus on important alignments when inferring the image-text similarity. It mainly focuses on local pair-wise matching between regions and words. In contrast, the proposed VSRN performs reasoning on region features and generate a global scene representation that captures key objects and semantic concepts for each image. This representation can be better aligned with the corresponding text caption. The comparison shows the strength of region relationship reasoning and global semantic reasoning for image-text matching. Especially for the challenging caption retrieval task, VSRN shows strong robustness to distractors with a huge improvement (relative 12.6\%).



\begin{table}
\centering
\scalebox{0.77}{
\begin{tabular}{l|ccc|ccc}
\hline
\multirow{2}{*}{Methods} & \multicolumn{3}{c|}{Caption Retrieval} & \multicolumn{3}{c}{Image Retrieval} \\
&  R@$1$ & R@$5$ & R@$10$ & R@$1$ & R@$5$ & R@$10$ \\      
\hline
\multicolumn{1}{l}{(R-CNN, AlexNet) } \\
DVSA$_{\rm{CVPR'15}}$ \cite{karpathy2015deep}  & 22.2 & 48.2 & 61.4 & 15.2  & 37.7 & 50.5\\
HMlstm$_{\rm{ICCV'17}}$ \cite{niu2017hierarchical}  & 38.1 & - & 76.5 & 27.7 & - & 68.8 \\
\hline
\multicolumn{1}{l}{(VGG)} \\
FV$_{\rm{CVPR'15}}$ \cite{klein2015associating}  & 35.0 & 62.0 & 73.8 & 25.0 & 52.7 & 66.0 \\
VQA$_{\rm{ECCV'16}}$ \cite{lin2016leveraging}  & 33.9 & 62.5 & 74.5 & 24.9 & 52.6 & 64.8 \\
SMlstm$_{\rm{CVPR'17}}$ \cite{huang2017instance}  & 42.5 & 71.9 & 81.5 & 30.2 & 60.4 & 72.3 \\
2WayN$_{\rm{CVPR'17}}$ \cite{eisenschtat2017linking}  & 49.8 & 67.5 & - & 36.0 & 55.6 & - \\
\hline
\multicolumn{1}{l}{(ResNet)} \\
RRF$_{\rm{ICCV'17}}$ \cite{liu2017learning}  & 47.6 & 77.4 & 87.1 & 35.4 & 68.3 & 79.9\\
VSE++$_{\rm{BMVC'18}}$ \cite{faghri2017vse}  & 52.9 & 79.1 & 87.2 & 39.6 & 69.6 & 79.5 \\
SCO$_{\rm{CVPR'18}}$ \cite{huang2018learning} & 55.5 & 82.0& 89.3 & 41.1 & 70.5 & 80.1 \\
\hline
\multicolumn{3}{l}{(Faster R-CNN, ResNet) } \\
SCAN$_{\rm{ECCV'18}}$ \cite{lee2018stacked}  & 67.4 & 90.3 & 95.8 & 48.6 & 77.7 & 85.2 \\  
VSRN (ours)  & \textbf{71.3} & \textbf{90.6} & \textbf{96.0} & \textbf{54.7} & \textbf{81.8} & \textbf{88.2}  \\
\hline
\end{tabular}}
\vspace{1mm}
\caption{Quantitative evaluation results of the image-to-text (caption) retrieval and text-to-image (image) retrieval on Fliker30K test set in terms of Recall@K (R@K).}
\label{tb:flicker_results}
\end{table}

\subsection{Ablation Studies}

\textbf{Analysis each reasoning component in VSRN.}
We would like to incrementally validate each reasoning component in our VSRN by starting from a very basic baseline model which does not perform any reasoning. This baseline model adopts a mean-pooling operation on the region features after the fully-connected layer $V = \{ {v_1},...,{v_k}\} ,{v_i} \in {\mathbb{R}^{{D}}}$ to obtain the final representation for the whole image ${I} \in {\mathbb{R}^{{D}}}$. The other parts are kept as the same as VSRN. Results on MS-COCO 1K test set are shown in Table~\ref{tb:ablation_model_coco}. This baseline model (noted as `Mean-pool') achieves 64.3 of R@1 for caption retrieval and 49.2 of R@1 for image retrieval. Then we add one region relationship reasoning (RRR) layer (described in Sec.~\ref{sc:global_context_R}) before the mean-pooling operation into this baseline model and mark it as RRR. We also replace the mean-pooling operation with the global semantic reasoning (GSR) module (described in Sec.~\ref{sc:global_context_R}) to get a GSR model. From Table~\ref{tb:ablation_model_coco} we can find that these two reasoning modules can both help to obtain better image representation ${I}$ and improve the matching performance effectively.

\begin{figure*}
\centering
\includegraphics[width=1.0\linewidth]{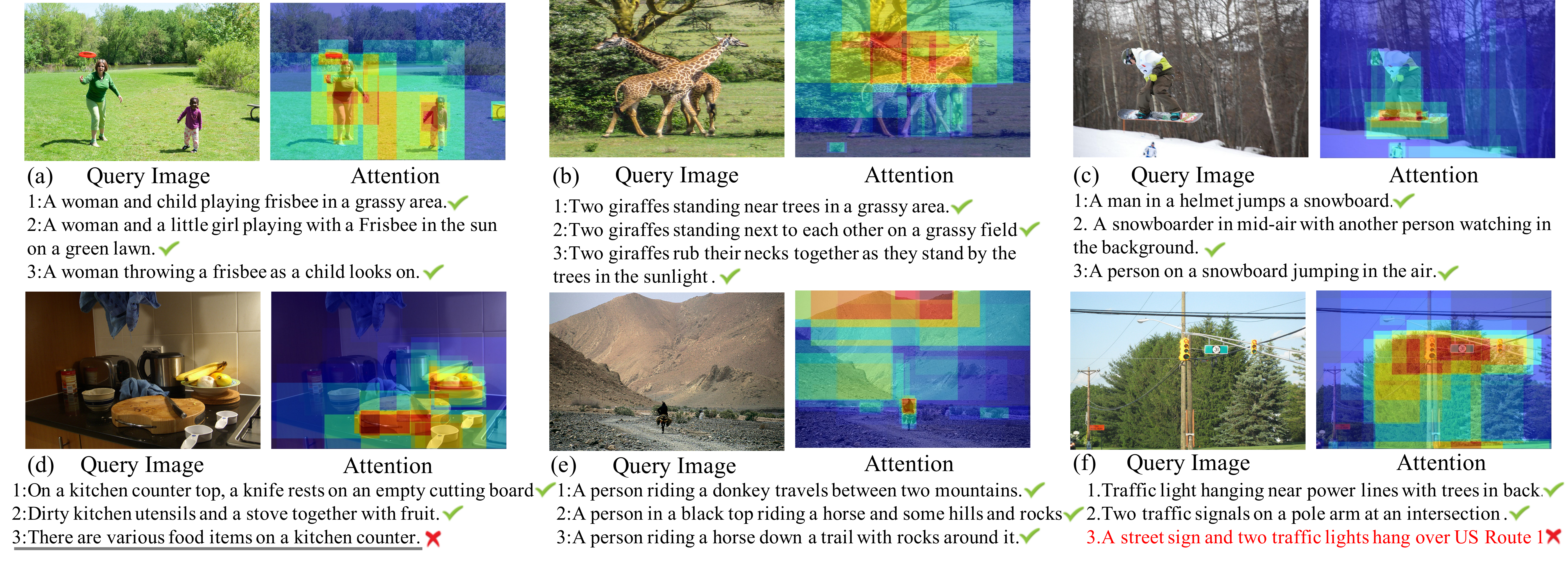} 
\caption{Qualitative results of the image-to-text (caption) retrieval for VSRN on MS-COCO dataset. For each image query, we show the top-3 ranked text caption. Ground-truth matched sentences are with check marks, while some sentences sharing similar meanings as ground-truth ones are marked with gray underline. We also show the attention visualization of the final image representation besides its corresponding image. Our model generates interpretable image representation that captures key objects and semantic concepts in the scene. (Best viewed in color when zoomed in.)}
\label{fig:Q_i2t} 
\end{figure*}

We then combine RRR and GSR to get our VSRN model and further try different numbers of RRR layers. Results show that adding region relationship reasoning layers before the global semantic reasoning module can gradually help to achieve better performance. This is because the RRR module can generate relationship enhanced features, which allows GSR perform reasoning on a graph topology and consider both current local region and global semantic correlations. However, we also find improvements become less when adding more RRR layers. We finally take 4RRR+GSR as the final setting of VSRN. We further report results of VSRN trained without text generation loss $L_{G}$ (marked as 4RRR+GSR*). Comparison shows that the joint optimization of matching and generation can help to improve around 2\% relatively for R@1. 

\begin{table}
\centering
\scalebox{0.84}{
\begin{tabular}{l|ccc|ccc}
\hline
\multirow{2}{*}{Methods} & \multicolumn{3}{c|}{Caption Retrieval} & \multicolumn{3}{c}{Image Retrieval} \\
&  R@$1$ & R@$5$ & R@$10$ & R@$1$ & R@$5$ & R@$10$ \\      
\hline
Mean-pool & 64.3 & 90.5 & 95.1 & 49.2 & 83.4 & 91.5 \\    
RRR  & 68.5 & 93.2 & 96.3 & 56.8 & 87.2 & 94.2 \\
GSR  & 72.3 & 94.4 & 98.0 & 59.6 & 88.6 & 94.5 \\  
1RRR + GSR  & 75.3 & 94.7 & 98.1 & 62.1 & 89.2 & 94.9 \\
4RRR + GSR & 76.2 & 94.8 & 98.2 & 62.8 & 89.7 & 95.1\\ 
4RRR + GSR* & 74.6 & 94.6 & 98.2 & 61.2 & 89.0 &  94.8\\

\hline
\end{tabular}}
\vspace{1mm}
\caption{Ablation studies on the MS-COCO 1K test set. Results are reported in terms of Recall@K (R@K). ``RRR'' means model with region relationship reasoning module. ``GSR'' represents a model with global semantic reasoning module. The number before RRR represents the number of RRR layers. ``*'' means model training without using text generation loss $L_{G}$. }
\label{tb:ablation_model_coco}
\end{table}

\begin{table}
\centering
\scalebox{0.80}{
\begin{tabular}{l|ccc|ccc}
\hline
\multirow{2}{*}{Methods} & \multicolumn{3}{c|}{Caption Retrieval} & \multicolumn{3}{c}{Image Retrieval} \\
&  R@$1$ & R@$5$ & R@$10$ & R@$1$ & R@$5$ & R@$10$ \\      
\hline
VSRN-Random  & 75.1 & 94.5 & 98.0 & 62.3 & 89.1 & 94.6  \\ 
VSRN-BboxSize  & 75.8 & 94.9 & 98.4 & 62.5 & 89.5 & 94.8 \\    
VSRN-Confidence &  76.2 & 94.8 & 98.2 & 62.8 & 89.7 & 95.1  \\ 
\hline
\end{tabular}}
\vspace{1mm}
\caption{Ablation studies on the MS-COCO 1K test set to analyze region ordering for global semantic reasoning. Results are reported in terms of Recall@K (R@K).}
\label{tb:ablation_ordering}
\end{table}

\textbf{Region ordering for global semantic reasoning.}
Since our global semantic reasoning module (Sec.~\ref{sc:global_context_R}) sequentially processes region features and generates the representation of the whole image gradually, we consider several ablations about region ordering for this reasoning process in Table~\ref{tb:ablation_ordering}. One possible setting (VSRN-Confidence) is the descending order of their class detection confidence scores that are generated by the bottom-up attention detector. We expect this to encourage the model to focus on the easy regions with high confidence first and then inferring more difficult regions based on the semantic context. Another option (VSRN-BboxSize) is to sort the detection bounding boxes of these regions in descending order, as this lets the model to obtain global scene information first. We also test the model with randomly ordering of the regions (VSRN-Random). Results in Table~\ref{tb:ablation_ordering} show that reasoning in a specific oder can help improve the performance than the random one. VSRN-Confidence and VSRN-BboxSize achieve comparable results with a reasonable ordering scheme. We take VSRN-Confidence as the setting of VSRN in our previous experiments. Besides, we also find the variance of R@1 is around 1 point for these different settings, which suggests VSRN is robust to the ordering scheme used. One possible reason could be that global information is included during the region relationship reasoning step. Based on these relationship enhanced feature, semantic reasoning can be then performed on global graph topologies.

\begin{figure*}
\centering
\includegraphics[width=0.98\linewidth]{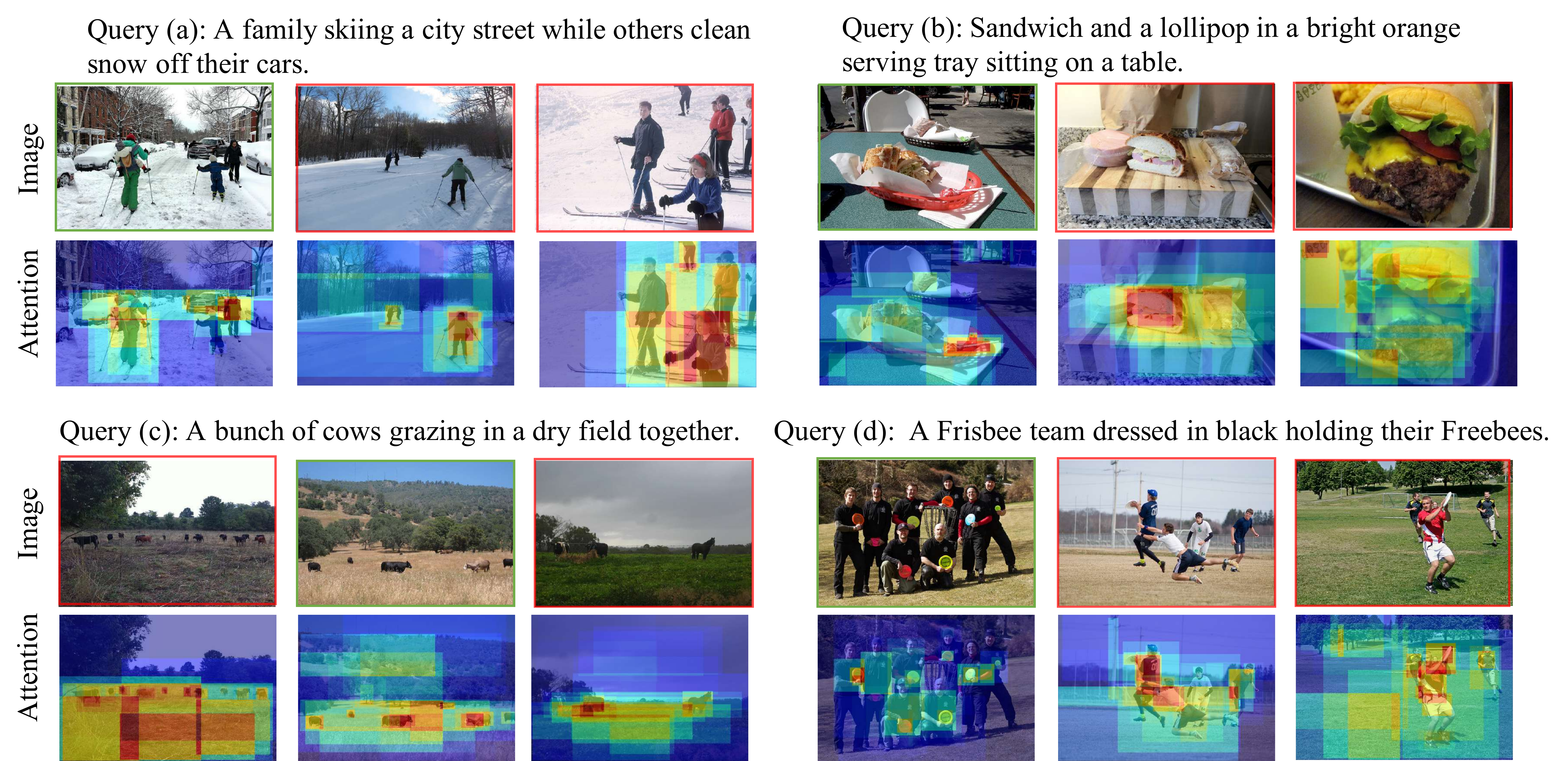} 
\caption{Qualitative results of the text-to-image (image) retrieval for VSRN on MS-COCO dataset. We show the top-3 retrieved images for each text query, ranking from left to right. The true matches are outlined in green boxes and false matches in red boxes. We also show the attention visualization of image representation generated by VSRN under the corresponding image.}
\label{fig:Q_t2i} 
\end{figure*}

\subsection{Visualization and Analysis}\label{sc:visualization}

\textbf{Attention visualization of the final image representation.} Since the final goal of our visual semantic reasoning is to generate the image representation that includes key object and semantic concepts in the scene. In order to validate this, we visualize the correlation between the final representation of the whole image and these image regions included in this image in an attention form. Specifically, we calculate the inner product similarity (same as in Eq.~\ref{eq:loss_match}) between each region feature $V^* = \{ {v_1^*},...,{v_k^*}\} ,{v_i^*} \in {\mathbb{R}^{{D}}}$ and the final whole image representation ${I} \in {\mathbb{R}^{{D}}}$. Then we rank the image regions $V^*$ in the descending order of their correlation with ${I}$ and assign a score ${s_{i}}$ to each ${v_i^*}$ according to its rank ${r_{i}}$. The score is calculate by ${s_{i}} =\lambda {(k - {r_{attn}})^2}$, where $k$ is the total number of regions, $\lambda$ is a parameter used to emphasize the high ranked regions. We set $\lambda = 50$ in our experiments. Then for the final attention map (similarity map), the attention score at each pixel location is obtained by adding up scores of all regions it belongs to. We show attention maps of each image along with the qualitative results of the image-to-text (caption) retrieval and text-to-image (image) retrieval.

\textbf{Qualitative results of the image-to-text retrieval.}
In Figure \ref{fig:Q_i2t}, we show the qualitative results of the text retrieval given image queries on MS-COCO. We show the top-3 retrieved sentences for each image query. The rank is obtained based the similarity scores predicted by VSRN. From these results, we find that our VSRN can retrieve the correct results in the top ranked sentences even for cases of cluttered and complex scenes. The model outputs some reasonable mismatches, e.g. (d)-3. There are incorrect results such as (f.4), which is possibly due to the too specific concept in the image (``US Route 1'') that the model could not identify. From the attention visualization, we can find the image representation generated by VSRN well captures key objects and semantic concepts in the scene.


\textbf{Qualitative results of the text-to-image retrieval.}
In Figure \ref{fig:Q_t2i}, we show qualitative results of image retrieval for given text queries on MS-COCO. Each text description is matched with a ground-truth image. We show the top-3 retrieved images for each text query. The true matches are outlined in green and false matches in red. We find that our model can retrieve the ground truth image in the top-3 list. Note that other results are also reasonable, which include the objects of the same category or same semantic concepts with the text descriptions. For those images with a very similar scene, our model can still distinguish them well and accurately retrieve the ground truth one at top-1 rank. This can be well explained from the attention map, e.g. for the given text query (a), the model attends on the cars on the street and the person cleaning a car in the ground-truth image to distinguish it with the other two images that are also about people skiing. However, for the top-2 retrieval images of query (c), the model is confused about the concept of ``try field''. It treats the field with less grass as a better match than the field with withered grass. This is possibly due to not enough training data for a complex concept.





\section{Conclusion}
In this paper, we present a simple and interpretable reasoning model VSRN to generate visual representation by region relationship reasoning and global semantic reasoning. The enhanced image representation captures key objects and semantic concepts of a scene, so that it can better align with the corresponding text caption. Extensive experiments on MS-COCO and Fliker30K datasets demonstrate the resulting model consistently outperforms the-state-of-the-art methods with a large margin for the image-text matching. Compared with the complicated attention-based aggregation from pairwise similarities among regions and words, we show that the classical ``image-text'' similarity measure still promising given enhanced whole image representation. We will further explore the effectiveness of reasoning modules in VSRN on other vision and language tasks.


\section{Acknowledgments}
This research is supported in part by the NSF IIS award 1651902 and U.S. Army Research Office Award W911NF-17-1-0367.


{\small
\bibliographystyle{ieee_fullname}
\balance
\bibliography{egbib}

\begin{thebibliography}{10}\itemsep=-1pt

\bibitem{anderson2018bottom}
Peter Anderson, Xiaodong He, Chris Buehler, Damien Teney, Mark Johnson, Stephen
  Gould, and Lei Zhang.
\newblock Bottom-up and top-down attention for image captioning and visual
  question answering.
\newblock In {\em CVPR}, 2018.

\bibitem{chen2018iterative}
Xinlei Chen, Li-Jia Li, Li Fei-Fei, and Abhinav Gupta.
\newblock Iterative visual reasoning beyond convolutions.
\newblock In {\em CVPR}, 2018.

\bibitem{chung2014empirical}
Junyoung Chung, Caglar Gulcehre, KyungHyun Cho, and Yoshua Bengio.
\newblock Empirical evaluation of gated recurrent neural networks on sequence
  modeling.
\newblock {\em arXiv}, 2014.

\bibitem{eisenschtat2017linking}
Aviv Eisenschtat and Lior Wolf.
\newblock Linking image and text with 2-way nets.
\newblock In {\em CVPR}, 2017.

\bibitem{faghri2017vse}
Fartash Faghri, David~J Fleet, Jamie~Ryan Kiros, and Sanja Fidler.
\newblock Vse++: Improving visual-semantic embeddings with hard negatives.
\newblock In {\em BMVC}, 2018.

\bibitem{frome2013devise}
Andrea Frome, Greg~S Corrado, Jon Shlens, Samy Bengio, Jeff Dean, Tomas
  Mikolov, et~al.
\newblock Devise: A deep visual-semantic embedding model.
\newblock In {\em NIPS}, 2013.

\bibitem{girshick2014rich}
Ross Girshick, Jeff Donahue, Trevor Darrell, and Jitendra Malik.
\newblock Rich feature hierarchies for accurate object detection and semantic
  segmentation.
\newblock In {\em CVPR}, 2014.

\bibitem{gu2018look}
Jiuxiang Gu, Jianfei Cai, Shafiq~R Joty, Li Niu, and Gang Wang.
\newblock Look, imagine and match: Improving textual-visual cross-modal
  retrieval with generative models.
\newblock In {\em CVPR}, 2018.

\bibitem{gu2019scene}
Jiuxiang Gu, Handong Zhao, Zhe Lin, Sheng Li, Jianfei Cai, and Mingyang Ling.
\newblock Scene graph generation with external knowledge and image
  reconstruction.
\newblock In {\em CVPR}, 2019.

\bibitem{he2016deep}
Kaiming He, Xiangyu Zhang, Shaoqing Ren, and Jian Sun.
\newblock Deep residual learning for image recognition.
\newblock In {\em CVPR}, 2016.

\bibitem{hobbs1993interpretation}
Jerry~R Hobbs, Mark~E Stickel, and Paul Martin.
\newblock Interpretation as abduction.
\newblock {\em Artificial intelligence}, 1993.

\bibitem{hochreiter1997long}
Sepp Hochreiter and J{\"u}rgen Schmidhuber.
\newblock Long short-term memory.
\newblock {\em Neural computation}, 1997.

\bibitem{huang2017instance}
Yan Huang, Wei Wang, and Liang Wang.
\newblock Instance-aware image and sentence matching with selective multimodal
  lstm.
\newblock In {\em CVPR}, 2017.

\bibitem{huang2018learning}
Yan Huang, Qi Wu, Chunfeng Song, and Liang Wang.
\newblock Learning semantic concepts and order for image and sentence matching.
\newblock In {\em CVPR}, 2018.

\bibitem{karpathy2015deep}
Andrej Karpathy and Li Fei-Fei.
\newblock Deep visual-semantic alignments for generating image descriptions.
\newblock In {\em CVPR}, 2015.

\bibitem{katsuki2014bottom}
Fumi Katsuki and Christos Constantinidis.
\newblock Bottom-up and top-down attention: different processes and overlapping
  neural systems.
\newblock {\em The Neuroscientist}, 2014.

\bibitem{kingma2014adam}
Diederik~P Kingma and Jimmy Ba.
\newblock Adam: A method for stochastic optimization.
\newblock {\em arXiv}, 2014.

\bibitem{kipf2017semi}
Thomas~N. Kipf and Max Welling.
\newblock Semi-supervised classification with graph convolutional networks.
\newblock In {\em ICLR}, 2017.

\bibitem{kiros2014unifying}
Ryan Kiros, Ruslan Salakhutdinov, and Richard~S Zemel.
\newblock Unifying visual-semantic embeddings with multimodal neural language
  models.
\newblock {\em arXiv}, 2014.

\bibitem{klein2015associating}
Benjamin Klein, Guy Lev, Gil Sadeh, and Lior Wolf.
\newblock Associating neural word embeddings with deep image representations
  using fisher vectors.
\newblock In {\em CVPR}, 2015.

\bibitem{krishna2017visual}
Ranjay Krishna, Yuke Zhu, Oliver Groth, Justin Johnson, Kenji Hata, Joshua
  Kravitz, Stephanie Chen, Yannis Kalantidis, Li-Jia Li, David~A Shamma, et~al.
\newblock Visual genome: Connecting language and vision using crowdsourced
  dense image annotations.
\newblock {\em IJCV}, 2017.

\bibitem{lao2011random}
Ni Lao, Tom Mitchell, and William~W Cohen.
\newblock Random walk inference and learning in a large scale knowledge base.
\newblock In {\em EMNLP}, 2011.

\bibitem{lee2018stacked}
Kuang-Huei Lee, Xi Chen, Gang Hua, Houdong Hu, and Xiaodong He.
\newblock Stacked cross attention for image-text matching.
\newblock In {\em ECCV}, 2018.

\bibitem{li2018support}
Kai Li, Zhengming Ding, Kunpeng Li, Yulun Zhang, and Yun Fu.
\newblock Support neighbor loss for person re-identification.
\newblock In {\em ACM Multimedia}, 2018.

\bibitem{li2018tell}
Kunpeng Li, Ziyan Wu, Kuan-Chuan Peng, Jan Ernst, and Yun Fu.
\newblock Tell me where to look: Guided attention inference network.
\newblock In {\em CVPR}, 2018.

\bibitem{li2019gain}
Kunpeng Li, Ziyan Wu, Kuan-Chuan Peng, Jan Ernst, and Yun Fu.
\newblock Guided attention inference network.
\newblock {\em TPAMI}, 2019.

\bibitem{li2019attnbn}
Kunpeng Li, Yulun Zhang, Kai Li, Yuanyuan Li, and Yun Fu.
\newblock Attention bridging network for knowledge transfer.
\newblock In {\em ICCV}, 2019.

\bibitem{lin2014microsoft}
Tsung-Yi Lin, Michael Maire, Serge Belongie, James Hays, Pietro Perona, Deva
  Ramanan, Piotr Doll{\'a}r, and C~Lawrence Zitnick.
\newblock Microsoft coco: Common objects in context.
\newblock In {\em ECCV}, 2014.

\bibitem{lin2016leveraging}
Xiao Lin and Devi Parikh.
\newblock Leveraging visual question answering for image-caption ranking.
\newblock In {\em ECCV}, 2016.

\bibitem{liu2017learning}
Yu Liu, Yanming Guo, Erwin~M Bakker, and Michael~S Lew.
\newblock Learning a recurrent residual fusion network for multimodal matching.
\newblock In {\em ICCV}, 2017.

\bibitem{newell1980physical}
Allen Newell.
\newblock Physical symbol systems.
\newblock {\em Cognitive science}, 1980.

\bibitem{niu2017hierarchical}
Zhenxing Niu, Mo Zhou, Le Wang, Xinbo Gao, and Gang Hua.
\newblock Hierarchical multimodal lstm for dense visual-semantic embedding.
\newblock In {\em ICCV}, 2017.

\bibitem{ren2015faster}
Shaoqing Ren, Kaiming He, Ross Girshick, and Jian Sun.
\newblock Faster r-cnn: Towards real-time object detection with region proposal
  networks.
\newblock In {\em NIPS}, 2015.

\bibitem{santoro2017simple}
Adam Santoro, David Raposo, David~G Barrett, Mateusz Malinowski, Razvan
  Pascanu, Peter Battaglia, and Timothy Lillicrap.
\newblock A simple neural network module for relational reasoning.
\newblock In {\em NIPS}, 2017.

\bibitem{vendrov2015order}
Ivan Vendrov, Ryan Kiros, Sanja Fidler, and Raquel Urtasun.
\newblock Order-embeddings of images and language.
\newblock In {\em ICLR}, 2016.

\bibitem{venugopalan2015sequence}
Subhashini Venugopalan, Marcus Rohrbach, Jeffrey Donahue, Raymond Mooney,
  Trevor Darrell, and Kate Saenko.
\newblock Sequence to sequence-video to text.
\newblock In {\em ICCV}, 2015.

\bibitem{yang2018visual}
Wei Yang, Xiaolong Wang, Ali Farhadi, Abhinav Gupta, and Roozbeh Mottaghi.
\newblock Visual semantic navigation using scene priors.
\newblock {\em ICLR}, 2019.

\bibitem{yao2018exploring}
Ting Yao, Yingwei Pan, Yehao Li, and Tao Mei.
\newblock Exploring visual relationship for image captioning.
\newblock In {\em ECCV}, 2018.

\bibitem{young2014image}
Peter Young, Alice Lai, Micah Hodosh, and Julia Hockenmaier.
\newblock From image descriptions to visual denotations: New similarity metrics
  for semantic inference over event descriptions.
\newblock {\em TACL}, 2014.

\bibitem{zhang2019mst}
Yulun Zhang, Chen Fang, Yilin Wang, Zhaowen Wang, Zhe Lin, Yun Fu, and Jimei
  Yang.
\newblock Multimodal style transfer via graph cuts.
\newblock In {\em ICCV}, 2019.

\bibitem{zhou2018temporal}
Bolei Zhou, Alex Andonian, Aude Oliva, and Antonio Torralba.
\newblock Temporal relational reasoning in videos.
\newblock In {\em ECCV}, 2018.

\end{thebibliography}
}

\end{document}